\documentclass[twoside,twocolumn]{article} 
\usepackage{ltexpprt}

\newif\ifapx
\newif\ifarxiv
\arxivtrue
\apxtrue 

\newcommand{\ourmaintitle}{We Are Not Your Real Parents:\\Telling Causal from Confounded using MDL}
\newcommand{\ourtitle}{We Are Not Your Real Parents: Telling Causal from Confounded using MDL}
\newcommand{\ourmethod}{\textsc{CoCa}\xspace}
\newcommand{\oururl}{\url{http://eda.mmci.uni-saarland.de/coca/}}
\newcommand{\codeurl}{\oururl}

\usepackage[pdftitle={\ourtitle}, pdfauthor={David Kaltenpoth and Jilles Vreeken}, hidelinks, hyperfootnotes = false]{hyperref}

\makeatletter
\newif\if@restonecol
\makeatother

\usepackage{etex}
\usepackage[ruled,linesnumbered,vlined]{algorithm2e}
\usepackage{microtype}
\usepackage[tt=false]{libertine}
\usepackage[varqu]{zi4}
\usepackage[libertine]{newtxmath}
\usepackage[notheorems]{eda} 

\usepackage{adjustbox}

\usepackage{graphicx}

\pgfplotsset{compat=newest}
\usepgfplotslibrary{colorbrewer} 
\usepgfplotslibrary{fillbetween}
\graphicspath{ {./figs/} }
\ifpdf 
\usetikzlibrary{external}
\fi
\SetKwComment{tcpas}{\{}{\}}
\SetCommentSty{textnormal}
\SetArgSty{textnormal}
\SetKw{False}{false}
\SetKw{True}{true}
\SetKw{Null}{null}
\SetKwInOut{Output}{output}
\SetKwInOut{Input}{input}
\SetKw{AND}{and}
\SetKw{OR}{or}
\SetKw{Continue}{continue}

\DeclareMathOperator*{\argmax}{argmax}

\DeclareMathOperator*{\N}{Normal}
\DeclareMathOperator*{\LN}{LogNormal}
\DeclareMathOperator*{\Lap}{Laplace}
\DeclareMathOperator*{\U}{Uniform}

\makeatletter
\renewcommand*{\@fnsymbol}[1]{\ensuremath{\ifcase#1\or   \circ\or \bullet\or *\or \ddagger\or
   \mathsection\or \mathparagraph\or \|\or **\or \dagger\dagger
   \or \ddagger\ddagger \else\@ctrerr\fi}}
\makeatother

\newcommand\independent{\protect\mathpalette{\protect\independenT}{\perp}}
\def\independenT#1#2{\mathrel{\rlap{$#1#2$}\mkern2mu{#1#2}}}

\newcommand{\Xb}{\mathbf{X}}
\newcommand{\Wb}{\mathbf{W}}
\newcommand{\Zb}{\mathbf{Z}}
\newcommand{\wb}{\mathbf{w}}
\newcommand{\Mc}{\mathcal{M}}
\newcommand{\Nc}{\mathcal{N}}
\newcommand{\tca}{\text{ca}}
\newcommand{\tco}{\text{co}}

\begin{document}
\setlength{\pdfpagewidth}{8.5in}
\setlength{\pdfpageheight}{11in}

\title{\ourmaintitle}

\author{David Kaltenpoth\thanks{\rule{0pt}{1.1em}Max Planck Institute for Informatics and Saarland University, Saarbr\"{u}cken, Germany.
\texttt{dkaltenpo@mpi-inf.mpg.de}} \and Jilles Vreeken\thanks{Helmholtz Center for Information Security and Max Planck Institute for Informatics, Saarbr\"{u}cken, Germany. \texttt{jv@cispa-helmholtz.de}}} 

\date{}

\maketitle
   
\begin{abstract}
{\small\baselineskip=9pt 
Given data over variables $(X_1,...,X_m, Y)$ we consider the problem of finding out whether $X$ jointly causes $Y$ or whether they are all confounded by an unobserved latent variable $Z$. To do so, we take an information-theoretic approach based on Kolmogorov complexity. In a nutshell, we follow the postulate that first encoding the true cause, and then the effects given that cause, results in a shorter description than any other encoding of the observed variables.

The ideal score is not computable, and hence we have to approximate it. We propose to do so using the Minimum Description Length (MDL) principle. We compare the MDL scores under the models where $X$ causes $Y$ and where there exists a latent variables $Z$ confounding both $X$ and $Y$ and show our scores are consistent. To find potential confounders we propose using latent factor modeling, in particular, probabilistic PCA (PPCA).

Empirical evaluation on both synthetic and real-world data shows that our method, \ourmethod, performs very well---even when the true generating process of the data is far from the assumptions made by the models we use. Moreover, it is robust as its accuracy goes hand in hand with its confidence.

}
\end{abstract}

\section{Introduction}
\label{sec:intro}

Causal inference from observational data, i.e. inferring cause and effect from data that was not collected through randomized controlled trials, is one of the most challenging and important problems in statistics~\cite{pearl2009causality}. 
One of the main assumptions in causal inference is that of \emph{causal sufficiency}. That is, to make sensible statements on the causal relationship between two statistically dependent random variables $X$ and $Y$, it is assumed that there exists no hidden confounder $Z$ that causes both $X$ and $Y$. In practice this assumption is often violated---we seldom know all factors that could be relevant, nor do we measure everything---and hence existing methods are prone to spurious inferences.

In this paper, we study the problem of inferring whether $X$ and $Y$ are causally related, or, are more likely jointly caused by an unobserved confounding variable $Z$. To do so, we build upon the algorithmic Markov condition (AMC)~\cite{janzing2010causal}. This  recent postulate states that the simplest---measured in terms of Kolmogorov complexity---factorization of the joint distribution coincides with the true causal model. Simply put, this means that if $Z$ causes both $X$ and $Y$ the complexity of the factorization according to this model, $K(P(Z)) + K(P(X|Z)) + K(P(Y|Z))$, will be lower than the complexity corresponding to the model where $X$ causes $Y$, $K(P(Z)) + K(P(X)) + K(P(Y|X))$. As we obviously do not have access to $P(Z)$, we propose to estimate it using latent factor modelling. Second, as Kolmogorov complexity is not computable, we use the Minimum Description Length (MDL) principle as a well-founded approach to approximate it from above. This is the method that we develop in this paper.

In particular, we consider the setting where given a sample over the joint distribution $P(\mathbf{X},Y)$ of continuous-valued univariate or multivariate random variable $\mathbf{X} = (X_1, \ldots, X_m)$, and a continuous-valued scalar $Y$. Although it has received little attention so far, we are not the first to study this problem. Recently, Janzing and Sch\"{o}lkopf~\cite{janzing2017detecting,janzing2018detecting} showed how to measure the ``structural strength of confounding'' for linear models using resp. spectral analysis~\cite{janzing2017detecting} and ICA~\cite{janzing2018detecting}. Rather than implicitly measuring the significance, we explicitly model the hidden confounder $Z$ via probabilistic PCA. While this means our approach is also linear in nature, it gives us the advantage that we can fairly compare the scores for the models $X \rightarrow Y$ and $X \leftarrow Z \rightarrow Y$, allowing us to define a reliable confidence measure. 

Through extensive empirical evaluation on synthetic and real-world data, we show that our method, \ourmethod, short for Confounded-or-Causal, performs well in practice. This includes settings where the modelling assumptions hold, but also in adversarial settings where they do not. We show that \ourmethod beats both baselines as well as the recent proposals mentioned above. Importantly, we observe that our confidence score strongly correlates with accuracy. That is, for those cases where we observe a large difference between the scores for causal resp. confounded, we can trust \ourmethod to provide highly accurate inferences. 

The main contributions of this paper are as follows, we
 \begin{itemize}[noitemsep,topsep=0pt]
 	\item[(a)] extend the AMC with latent factor models, and propose to instantiate it via probabilistic PCA,
 	\item[(b)] define a consistent and easily computable MDL-score to instantiate the framework in practice,
 	\item[(c)] provide extensive evaluation on synthetic and real data, including comparisons to the state-of-the-art.
 \end{itemize}

This paper is structured as usual. In Sec.~\ref{sec:prelim} we introduce basic concepts of causal inference, and hidden confounders. We formalize our information theoretic approach to inferring causal or confounded in Sec.~\ref{sec:theory}. We discuss related work in Sec.~\ref{sec:related}, and present the experiments in Sec.~\ref{sec:exps}. Finally, we wrap up with discussion and conclusions in Sec.~\ref{sec:discussion}.

\newtheorem{definition}{Definition}

\section{Causal Inference and Confounding}
\label{sec:prelim}

In this work, we consider the setting where we are given $n$ samples from the joint distribution $P(\Xb,Y)$ over two statistically dependent continuous-valued random variables $\Xb$ and $Y$. We require $Y$ to be a scalar, i.e. univariate, but allow $\Xb = (X_1,\ldots,X_m)$ to be of arbitrary dimensionality, i.e. univariate or multivariate. Our task is to determine whether it is more likely that $\Xb$ jointly cause $Y$, or that there exists an unobserved random variable $\Zb = (Z_1,\ldots,Z_k)$ that is the cause of both $\Xb$ and $Y$. Before we detail our approach, we introduce some basic notions, and explain why the straightforward solution does not work.

\subsection{Basic Setup}

It is impossible to do causal inference from observational data without making assumptions~\cite{pearl2009causality}. That is, we can only reason about what we should observe in the data if we were to change the causal model, if we assume (properties of) a causal model in the first place. 

A core assumption in causal inference is that the data was drawn from a probabilistic graphical model, a casual directed acyclic graph (DAG). To have a fighting chance to recover this causal graph $G$ from observational data, we have to make two further assumptions. The first, and in practice most troublesome is that of \emph{causal sufficiency}. This assumption is satisfied if we have measured \emph{all} common causes of \emph{all} measured variables. This is related to Reichenbach's principle of common cause~\cite{reichenbach1956direction}, which states that if we find that two random variables $X$ and $Y$ are statistically dependent, denoted as $X \not \! \independent Y$, there are three possible explanations. Either $X$ causes $Y$, $X\rightarrow Y$, or, the other way around, $Y$ causes $X$, $X \leftarrow Y$, or there is a third variable $Z$ that causes both $X$ and $Y$, $X\leftarrow Z \rightarrow Y$. In order to determine the latter case, we need to have measured $Z$. 

The second additional assumption we have to make is that of \emph{faithfulness}, which is defined as follows. 
\begin{definition}[Faithfulness]
If a Bayesian network $G$ is faithful to a probability distribution $P$, then for each pair of nodes $X_i$ and $X_j$ in $G$, $X_i$ and $X_j$ are adjacent in $G$ iff. $X_i \not \! \independent X_j \mid \textbf{Z}$, for each $\textbf{Z} \subset G$, with $X_i, X_j \not \in \textbf{Z}$.
\end{definition}
In other words, if we measure that $X$ is independent of $Y$, denoted as $X \independent Y$, there is no direct influence between the two in the underlying causal graph. This is a strong, but generally reasonable assumption; after all, violations of this condition do generally not occur unless the distributions have been specifically chosen to this end.

Under these assumptions, Pearl~\cite{pearl2009causality} showed that we can factorize the joint distribution over the measured variables,
\begin{equation}
P(X_1, \ldots, X_m) = \prod_{i=1}^m P(X_i \mid \mathit{PA}_i) \; . \label{eq:markov}
\end{equation}
That is, we can write it as a product of the marginal distributions of each $X_i$ conditioned on its true causal parents $\mathit{PA}_i$. This is referred to as the causal Markov condition, and implies that, under all of the above assumptions, we can have a hope of reconstructing the causal graph from a sample from the joint distribution.

\subsection{Crude Solutions That Do Not Work}
\label{sec:problems}

Based on the above, many methods have been proposed to infer causal relationships from a dataset. We give a high level overview of the state of the art in Sec.~\ref{sec:related}. Here, we continue to discuss why it is difficult to determine whether a given pair $\Xb,Y$ is confounded or not, and in particular, why traditional approaches based on probability theory or (conditional) independence, do not suffice. 

To see this, let us first suppose that $\Xb$ causes $Y$, and there are is no hidden confounder $\Zb$. We then have $P(\Xb,Y) = P(\Xb)P(Y|\Xb)$, and $\Xb \not\!\independent Y$. Now, let us suppose instead that $\Zb$ causes $\Xb$ and $Y$, i.e.  $\Xb\leftarrow \Zb \rightarrow Y$. Then we would have $P(\Xb,Y,\Zb) = P(\Zb) P(\Xb|Z) P(Y| \Zb)$, and, importantly, while $X \independent Y \mid \Zb$, we still observe $X \not\!\independent Y$, and hence cannot determine causal or confounded on that alone. 

Moreover, as we are only given a sample over $P(\Xb,Y)$ for which $X \not\!\independent Y$ holds, but know nothing about $\Zb$ or $P(\Zb)$, we cannot directly measure $\Xb \independent Y \mid \Zb$. A simple approach would be to see if we can \emph{generate} a $\hat{\Zb}$ such that $\Xb \independent Y \mid \hat{\Zb}$; for example through sampling or optimization. However, as we have to assign $n$ values for $\hat{\Zb}$, this means we have $n$ degrees of freedom, and it easy to see that under these conditions it is \emph{always} possible to generate a $\hat{\Zb}$ that achieves this independence, even when there was no confounding $\Zb$. A trivial example is to simply set $\hat{\Zb} = \Xb$. 

A similarly flawed idea would be to decide on the likelihoods of the data alone, i.e. to see if we can find a $\hat{\Zb}$ for which $P(\hat{\Zb})P(\Xb|\hat{\Zb})P(Y|\hat{\Zb}) > P(\hat{\Zb})P(\Xb)P(Y|\Xb)$. Besides having to choose a prior on $\Zb$, as we already achieve equality by initializing $\hat{\Zb} = \Xb$ and have $n$ degrees of freedom, we again virtually always will find a $\hat{\Zb}$ for which this holds, regardless of whether there was a true confounder or not.

Essentially, the problem here is that it is too easy to find a $\hat{\Zb}$ where these conditions hold, which for a large part is due to the fact that we do not take the complexity of $\hat{\Zb}$ into account, and hence face the problem of overfitting. To avoid this, we take an information theoretic approach, such that in principled manner we can take both the complexity of $\hat{\Zb}$, as well as its effect on $\Xb$ and $Y$ into account. 

%
%


\section{Telling Causal from Confounded by Simplicity}
\label{sec:theory}

We base our approach on the algorithmic Markov condition, which in turn is based on the notion of Kolmogorov complexity. We first give short introductions to both notions, and then develop our approach.

\subsection{Kolmogorov Complexity} The Kolmogorov complexity of a finite binary string $x$ is the length of the shortest program $p^*$ for a universal Turing machine $\mathcal{U}$ that generates $x$ and then halts~\cite{li2009introduction,kolmogorov:63:random}. Formally, 
\begin{align}
  K(x) = \min \left\{ \left| p \right| : p \in \left\{ 0, 1 \right\}^{\ast}, \mathcal{U}(p) = x \right\} \; .
\end{align}
That is, program $p^*$ is the most succinct \emph{algorithmic} description of $x$, or, in other words, the ultimate lossless compressor of that string. 
For our purpose, we are particularly interested in the Kolmogorov complexity of a distribution $P$, 
\begin{align}
K(P) = \min \left\{ \left| p \right| : p \in \left\{ 0, 1 \right\}^{\ast}, \left| \mathcal{U}(x, p, q) - P(x) \right| \leq 1/q \right\} \; ,
\end{align}
which is the length of the shortest program $p^*$ for a universal Turing machine $\mathcal{U}$ that approximates $P$ arbitrarily well~\cite{li2009introduction}. 

By definition, Kolmogorov complexity will make maximal use of any structure in the input that can be used to compress the object. As such it is the theoretically optimal measure for complexity. Due to the halting problem, Kolmogorov complexity is also not computable, nor approximable up to arbitrary precision~\cite{li2009introduction}. The Minimum Description Length (MDL) principle~\cite{grunwald2007minimum}, however, provides a statistically well-founded approach to approximate it from above. We will later use MDL to instantiate the framework we define below. 

\subsection{Algorithmic Markov Condition}
\label{sec:amc}

Recently, Janzing and Sch\"{o}lkopf~\cite{janzing2010causal} postulated the \emph{algorithmic} Markov condition (AMC), which states that if $X$ causes $Y$, the factorization of the joint distribution over $X$ and $Y$ in the true causal direction has a lower Kolmogorov complexity than in the anti-causal direction, i.e. 
\begin{equation}
K(P(X)) + K(P(Y|X)) \leq K(P(Y))+K(P(X|Y)) \; 
\end{equation}
holds up to an additive constant.
Moreover, under the assumption of causal sufficiency this allows us to identify the true causal network as the least complex one,
\begin{align}
  K(P(X_1, \ldots, X_m)) = \min_G \sum_{i=1}^{m}K(P(X_i | \mathit{PA}_i)) \; , \label{eq:test} 
\end{align}
which again holds up to an additive constant.

%


\subsection{AMC and Confounding}
\label{sec:amcconf}

Although the algorithmic Markov condition relies on causal sufficiency, it does suggest a powerful inference framework where we do allow variables to be unobserved.
For simplicity of notation, as well as generality, let us ignore $Y$ for now, and instead consider the question whether $\Xb$ is confounded by some factor $\Zb$. We can answer this question using the AMC by including a latent variable $\Zb = (Z_1,...,Z_k)$, where we assume the $Z_j$'s to be independent, of which we know the joint distribution corresponding to measured $\Xb$ and unmeasured $\Zb$, $P(\Xb, \Zb)$. If this is the case, we can again simply identify the corresponding minimal Kolmogorov complexity network via
\begin{align}
  K(P(\Xb,\Zb)) = \min_{G} \sum_{i=1}^m K(P(X_i | \mathit{PA}_i)) + \sum_{j=1}^k K(P(Z_j))\, , \label{eq:amcc}
\end{align}
where $\mathit{PA}_i$ are now the parents of $X_i$ among $\left\{ X_l, Z_j \right\}$ in the extended network. By adding terms $K(P(Z_j))$ we implicitly assume that there is no reverse causality $\Xb \rightarrow \Zb$. 

This formulation gives us a principled manner to identify whether a given $P(\Zb)$ is a (likely) confounder of $\Xb$. Clearly, with the above we can score the hypothesis $\Zb \rightarrow \Xb$. However, it also allows us to fairly score the hypothesis $\Zb \independent \Xb$, because if we choose $P(\Zb)$ to be a prior concentrated on a single point, this corresponds to Eq.~\eqref{eq:test} up to an additive constant. By the algorithmic Markov condition, we can now determine the most likely causal model, simply by comparing the two scores and choosing the one with the lower Kolmogorov complexity. This approach does not suffer from the same problems as in Sec.~\ref{sec:problems} as we explicitly take the complexity of $P(\Zb)$ into account.
Moreover, and importantly, this formulation allows us to consider \emph{any} distribution $P(\Xb,\Zb)$ with \emph{any} type of latent factor $\Zb$. 


Two problems, however, do remain with this approach. First, we do not know the true distribution $P(\Xb, \Zb)$, nor even distributions $P(\Xb)$ or $P(\Zb)$. Instead we only have empirical data over $\Xb$ from which we can approximate $\hat{P}(\Xb)$, but this does not give us explicit information about $\Zb$, $P(\Zb)$ or the joint $P(\Xb, \Zb)$. Second, as stated above, Kolmogorov complexity is not computable and the criterion as such therefore not directly applicable. 
We will deal with the first problem next by making assumptions on the form of $P(\Xb, \Zb)$, and then in Sec.~\ref{sec:mdl} will instantiate this criterion using the Minimum Description Length (MDL) principle. 

\subsection{Latent Factor Models}
\label{sec:factor}

Even under the assumption that the $Z_j$ are mutually independent, there are infinitely many possible distributions $P(\Xb,\Zb)$, and hence we have to make further choices to make the problem feasible. In our setting, a particularly natural choice is to use latent factor modelling. That is, we say the distribution over $\Xb, \Zb$ should be of the form
\begin{align}\nonumber
  P(\Xb, \Zb) = P(\Zb)\prod_{i=1}^m P(X_i | \Zb)
\end{align}
where the distribution of $\Zb$ can be arbitrarily complex. Not only does this give us a very clear and interpretable hypothesis, namely that given $P(\Zb)$, every $X_i$ should be independent of every other member of $\Xb$, i.e. $X_i \independent X_j \mid \Zb$, it also corresponds to the notion that $P(\Zb)$ should explain away \emph{as much} of the information shared within $\Xb$ as possible---very much in line with Eq.~\eqref{eq:amcc}. Moreover, from a more practical perspective, it is also a well-studied problem for which advanced techniques exist, such as Factor Analysis \cite{loehlin1998latent}, 
GPLVM \cite{lawrence2005probabilistic}, Deep Generative Models \cite{kingma2013autoencoding,rezende2015variational,ranganath2015deep}, as well as Probabilistic PCA (PPCA) \cite{tipping1999probabilistic}. 

For the sake of simplicity we will here focus on using PPCA, which has the following linear form
\begin{align}
  Z_i &\sim \N(0, \sigma_z^2 I)  \label{eq:ppca} \\
  W_i &\sim \N(0, \sigma_w^2I) \nonumber \\
  \Xb | \Zb, \Wb &\sim \N(\Wb^t\Zb, \sigma_x^2I) \; , \nonumber
\end{align}
and is appropriate if we deal with real-valued variables without any constraints and assume Gaussian noise. If the data does not follow these assumptions one of the other models mentioned above may be a more appropriate choice. An appealing aspect of PPCA is that  by marginalizing over $\Zb$ we can rewrite it in only terms of the matrix $W$ \cite{tipping1999probabilistic}, i.e. 
\begin{align}
\label{eq:ppca-simp}
  W_i &\sim \N(0, \sigma_w^2I) \\
  \Xb | \Wb &\sim \N(0, \Wb \Wb^t + \sigma_x^2I) \; ,
\end{align}
which both dramatically reduces the computational effort as well as will allow us to make statements about the consistency of our method.

While in the simple form PPCA assumes linear relationships, we can also model non-linear relationships by adding features to conditional distribution $\Xb | \Zb, \Wb$, e.g. using polynomial regression of $\Xb$ on $\Zb$. While this increases the modelling power, it comes with an increase in computational effort as the simplification of Eq.~\eqref{eq:ppca-simp} no longer holds.

\subsection{Minimum Description Length}
\label{sec:mdl}

While Kolmogorov complexity is not computable, the Minimum Description Length (MDL) principle~\cite{rissanen:78:mdl} provides a statistically well-founded approach to approximate $K(\cdot)$ from above. To achieve this, rather than considering all Turing machines, in MDL we consider a model class $\Mc$ for which we \emph{know} that every model $M \in \Mc$ will generate the data and halt, and identify the best model $M^* \in \Mc$ as the one that describes the data most succinctly without loss. If we instantiate $\Mc$ with all Turing machines that do so, the MDL-optimal model coincides with Kolmogorov complexity---this is also known as Ideal MDL~\cite{grunwald2007minimum}. In practice, we of course consider smaller model classes that are easier to handle and match our modelling assumptions. 

In two-part, or, \emph{crude} MDL, we score models $M \in \Mc$ by first encoding the model, and then the data given that model, 
\begin{align}
L(X,M) = L(M) + L(X \mid M) \; ,
\end{align}
where $L(M)$ and $L(X | M)$ are code length functions for the model, and the data conditional on the model, respectively.

Two-part MDL often works well in practice, but, by encoding the model separately it introduces arbitrary choices. In one part MDL---also known as \emph{refined} MDL---we avoid these choices by encoding the data using the entire model class at once. In order for a code length function to be refined, it has to be asymptotically mini-max optimal. That is, no matter what data $X'$ of the same type and dimensions as $X$ we consider, the refined score for $X'$ is within a constant from the score where we already know its corresponding optimal model $M'^{*}$, $L(X' \mid M'^{*})$, and this constant is independent of the data. There exist different forms of refined MDL codes~\cite{grunwald2007minimum}. For our setup it is convenient to use the full Bayesian definition, 
\begin{align*}
  L(X | \Mc) = -\log \int_{M \in \Mc} P(X | M)dP(M)
\end{align*}
where $P(M)$ is a prior on the model class $\Mc$. 
In our case, that is, for the PPCA model from \ref{eq:ppca} each pair $\Zb, \Wb$ corresponds to one model $M$, and hence the model class to all possible $\Zb, \Wb$ of which the posterior is given by Eq.~\eqref{eq:ppca}, i.e. we have
\begin{align}
L(\Xb | \Mc) = -\log \int p(\Xb | \Zb, \Wb)p(\Zb)p(\Wb) d\Wb d\Zb \; .
\end{align}
We can now put all the pieces together, and use the above theory to determine whether a pair $\Xb,Y$ is more likely causally related or confounded by an unobserved $\Zb$.

\subsection{Causal or Confounded?}
\label{sec:caus}


Given the above theory, determining which of $\Xb \rightarrow Y$ and $\Xb \leftarrow \Zb \rightarrow Y$ is more likely, is fairly straightforward. To do so, we consider two model classes, one for each of these two hypotheses, and determine which of the two leads to the most succinct description of the data sample over $\Xb$ and $Y$. 

First, we consider the causal model class $\Mc_{\text{ca}}$ that consists of models where $\Xb$ causes $Y$ in linear fashion, 
\begin{align}
  X_i &\sim \N(0, \sigma_x^2I) \label{eq:dcaus} \\
  \wb &\sim \N(0, \sigma_w^2I) \nonumber \\
  Y | \Xb, \wb &\sim \N(\wb^t \Xb, \sigma_y^2) \;. \nonumber 
\end{align}
and writing $L_{\text{ca}}$ instead of $L(\cdot | \mathcal{M}_{\text{ca}})$ we encode the data as
\begin{align}
  L_{\tca}(\Xb, Y) &= -\log P(\Xb) \int P(Y \mid \Xb, \wb)P(\wb)d \wb \\
                                   &\approx -\log P(\Xb) N^{-1}\sum_{j = 1}^N P(Y | \Xb, \hat{\wb}_j)\label{eq:scaus}
\end{align}
where we approximate the integral by sampling $N$ weight vectors $\hat{\wb}_i$ from the distribution defined by Eq.~\eqref{eq:dcaus}.

Second, we consider the \emph{confounded} model class $\Mc_{\text{co}}$, where the correlations within $\Xb$ and $Y$ are entirely explained by a hidden confounder modelled by PPCA, i.e.
\begin{align}\label{eq:conf}
  L_{\tco}(\Xb, Y) &= -\log \int p(\Xb, Y | \Zb, \Wb)p(Z)p(\Wb) d\Wb d\Zb \\
                                   &\approx -\log N^{-1} \sum_{j = 1}^N p(\Xb, Y | \hat{\Zb}_j, \hat{\Wb}_j)
\end{align}
where the $N$ samples for $\hat{\Zb}_j, \hat{\Wb}_j$ are drawn from the model we inferred using PPCA, i.e., according to Eq.~\eqref{eq:ppca}. Like for the causal case, the more samples we consider, the better the approximation, but the higher the computational cost. 

By MDL we can now check which hypothesis better explains the data, by simply considering the sign of $L_{\tco}(\Xb, Y) - L_{\tca}(\Xb, Y)$.
If this is less than zero, the confounded model does a better job at describing the data than the causal model and vice versa. We refer to this approach as \ourmethod.

To make the \ourmethod scores comparable between different data sets, we further introduce the confidence score
\begin{align*}
  C = \frac{L(\Xb, Y | \Mc_{\text{co}}) - L(\Xb, Y | \Mc_{\text{ca}})}{\max \left\{ L(\Xb, Y | \Mc_{\text{co}}) , L(\Xb, Y | \Mc_{\text{ca}}) \right\}} \; ,
\end{align*}
which is simply a normalized version of $c$ that accounts for both the intrinsic complexities of the data as well as the number of samples. If the absolute value of $C$ is small both model classes explain the data approximately equally well, and hence we are not very confident in our result and should perhaps refrain from making a decision. 



Last, we consider the question of whether we can say when our method will properly distinguish between the cases we care about? For this, we use a general result for MDL on the consistency of deciding between two model classes when the data is generated by a model contained in either of these classes~\cite{grunwald2007minimum}. That is, if we let $\Xb^n, Y^n$ be $n$ samples for $\Xb$ and $Y$ then
\begin{align}
\label{eq:consistency}
  \lim_{n \rightarrow \infty} & n^{-1} \left( L_{\tco}(\Xb^n, Y^n) - L(\Xb^n, Y^n) \right)
  \left\{
  \begin{array}{lr}
    \!\leq 0 & \text{if } M^{\ast} \in \Mc_{\text{co}}  \\
    \!\geq 0 & \text{if } M^{\ast} \in \Mc_{\text{ca}}
  \end{array}
\right.
\end{align}
with strict inequalities if $M^{\ast}$ is contained in only one of the two classes. This means that in the limit we will infer the correct conclusion if the true model is within the model classes we assume. Moreover, since our refined MDL formulation is also consistent for model selection~\cite{grunwald2007minimum}, following Sec.~\ref{sec:amc} we expect that even if $M^{\ast}$ is contained in both model classes the shortest description of the model $M^{\ast}$ corresponds to the true generative process. Importantly, even when the true model is not in either of our model classes, we can still expect reasonable inferences with regard to these model classes; by the minimax property of refined codes we use, we encode every model as efficiently (up to a constant) as possible, which promises reliable performance and confidence scores even in adversarial cases. As we will see shortly, the experiments confirm this.



\section{Related Work}
\label{sec:related}

Causal inference is arguably one of the most important problems in statistical inference, and hence has attracted a lot of research attention \cite{rubin1974estimating,pearl2009causality,spirtes2000causation}. The existence of confounders, selection bias and other statistical problems make it impossible to infer causality from observational data alone~\cite{pearl2009causality}. When their assumptions hold, constraint-based~\cite{spirtes2000causation,spirtes1999algorithm,zhang2008completeness} and score-based~\cite{chickering2002learning} causal discovery can, however, reconstruct causal graphs up to Markov equivalence. This means, however, they are not applicable to determine the causal direction between just $X$ and $Y$.


By making assumptions on the shape of the causal process Additive Noise Models (ANMs) can determine the causal direction between just $X$ and $Y$. In particular, ANMs assume independence between the cause and the residual (noise), and infer causation if such a model can be found in one direction but not in the other~\cite{shimizu2006linear,shimizu2011directlingam,hoyer2009nonlinear,zhang2009identifiability}. 
A more general framework for inferring causation than any of the above is given by the Algorithmic Markov Condition (AMC) \cite{lemeire2006causal,janzing2010causal} which is based on finding the least complex -- in terms of Kolmogorov complexity -- causal network for the data at hand. Since Kolmogorov complexity is not computable \cite{li2009introduction}, practical instantiations require a computable criterion to judge the complexity of a network, which has been proposed to do using Renyi-entropies~\cite{kocaoglu2016entropic},
information geometry~\cite{daniusis2012inferring,janzing2012informationgeometric,janzing2014justifying}, and MDL~\cite{budhathoki2017causal,marx2017causal}. All of these methods assume causal sufficiency, however, and are not applicable in the case where there are hidden confounders.


Rather than inferring the causal direction between $X$ and $Y$, estimating the causal effect of $X$ onto $Y$ is also an active topic of research. To do so in the presence of latent variables, Hoyer et al.~\cite{hoyer2008estimation} solve the overcomplete independent component analysis (ICA) problem, whereas Wang and Blei~\cite{wang2018blessings} and Ranganath and Perotte~\cite{ranganath2018multiple} control for plausible confounders using a given factor model. 

Most relevant to this paper is the recent work by Janzing and Sch\"{o}lkopf on determining the ``structural strength of confounding'' for a continuous-valued pair $\Xb, Y$, which they propose to measure using resp. spectral analysis \cite{janzing2017detecting} and ICA \cite{janzing2018detecting}. Like us, they also focus on linear relationships, but in contrast to us define a one-sided significance score, rather than a two-sided information theoretic confidence score. In the experiments we will compare to these two methods.


\section{Experiments}
\label{sec:exps}

In this section we empirically evaluate \ourmethod. In particular, we consider performance in telling causal from confounded for both in-model and adversarial settings on both synthetic and real-world data. We compare to the recent methods by Janzing and Sch\"{o}lkopf~\cite{janzing2017detecting,janzing2018detecting}.
We implemented \ourmethod in Python using PyMC3 \cite{salvatier2016probabilistic} for posterior inference via ADVI \cite{kucukelbir2017automatic}. All code is available for research purposes.\!\footnote{\codeurl} 

Throughout this section we infer one-dimensional factor models $\hat{\Zb}$, noting that higher-dimensional $\hat{\Zb}$ gave similar results. We use $N$$=$$500$ samples to calculate the MDL scores.
All experiments were executed single-threaded on an Intex Xeon E5-2643 v3 machine with 256GB memory running Linux, and each run took on the order of seconds to finish.

\subsection{Synthetic Data}
\label{sec:synth}

To see whether \ourmethod works at all, we start by generating synthetic data with known ground truth close to our assumptions. For the confounded case, we generate samples over $\Xb, Y$ as follows
\begin{align*}
  &Z_j \sim p_z, \; &W_{ij} \sim p_w \\
  &\epsilon \sim \N(0, 1) \; &\Xb, Y = \Wb^t\Zb + \epsilon \, ,
\end{align*}
while for the causal case, we generate $\Xb, Y$ as
\begin{align*}
  &X_i \sim p_x \; &w_i \sim p_w  \\
  &\epsilon \sim \N(0, 1) \; &Y = \wb^t\Xb + \epsilon \, .
\end{align*}
To see how our performance depends on the precise generating process, we consider the following source distributions,
\begin{align*}
  p_z, p_x, p_w \in \{ &\N(0, 1), \Lap(0, 1), \\
                       &\LN(0, 1), \U(0, 1) \} \, .
\end{align*}
We expect best \ourmethod performance when the generating process uses the $\N$ or $\Lap$ distributions as these are closest to the assumptions made in Eq.~\eqref{eq:ppca} and Eq.~\eqref{eq:dcaus}.

To see how the accuracy of \ourmethod depends on the confidence assigned to each inference, we consider decision rate (DR) plots.
In these, we consider the accuracy over the top-$k$ pairs sorted descending by absolute confidence, $|C|$.
This metric is commonly used in the literature on causal inference as it gives more information about the performance of our classifier than simple accuracy scores.

\begin{figure}[t]
  \ifarxiv
  \includegraphics{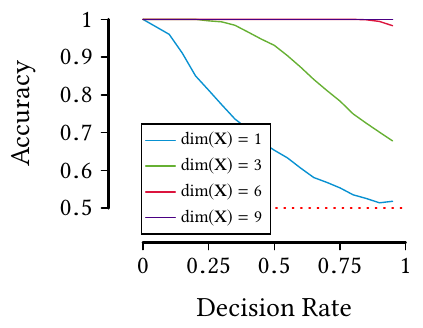}
  \includegraphics{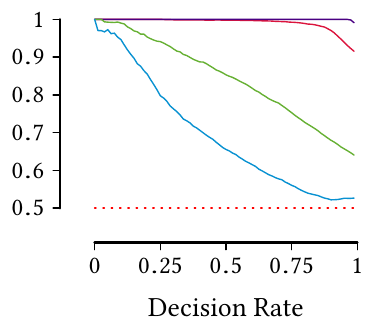}
  \else
  \begin{tikzpicture}
    \begin{axis}[eda line, ymax = 1.0, ymin = 0.5, ytick={1.0,0.9,0.8,0.7,0.6,0.5},
      ylabel={Accuracy}, xlabel={Decision Rate},
      ylabel near ticks, xlabel near ticks,
      tick label style={font=\footnotesize},
      label style={font=\footnotesize},
      height=3.5cm, width=4.25cm,
      legend pos=south west, legend columns=1,
      legend style={nodes={scale=0.7, transform shape}, at={(0.24,0.45)},anchor=north},
      xmin=0, xmax=1, xtick={0,0.25,0.5,0.75,1}]

      \addplot+[mambacolor4] table[x index = 0, y index = 1, header = false,
      each nth point=100] {expres/dr/CC_n_dx1_dz3.dat};
      \addlegendentry{$\dim(\Xb) = 1$}
      \addplot+[mambacolor3] table[x index = 0, y index = 1, header = false,
      each nth point=100] {expres/dr/CC_n_dx3_dz3.dat};
      \addlegendentry{$\dim(\Xb) = 3$}
      \addplot+[mambacolor6] table[x index = 0, y index = 1, header = false,
      each nth point=100] {expres/dr/CC_n_dx6_dz3.dat};
      \addlegendentry{$\dim(\Xb) = 6$}
      \addplot+[mambacolor1] table[x index = 0, y index = 1, header = false,
      each nth point=100] {expres/dr/CC_n_dx9_dz3.dat};
      \addlegendentry{$\dim(\Xb) = 9$}
      \addplot[dotted, mark=none, red] coordinates {(0,0.5) (1,0.5)};      
    \end{axis}
  \end{tikzpicture}%
  \begin{tikzpicture}
    \begin{axis}[eda line, ymax = 1.0, ymin = 0.5, ytick={1.0,0.9,0.8,0.7,0.6,0.5},
      ylabel={}, xlabel={Decision Rate},
      ylabel near ticks, xlabel near ticks,
      tick label style={font=\footnotesize},
      label style={font=\footnotesize},
      height=3.5cm, width=4.25cm,
      legend pos=south west, legend columns=1,
      legend style={nodes={scale=0.7, transform shape}, at={(0.25,0.5)},anchor=north},
      xmin=0, xmax=1, xtick={0,0.25,0.5,0.75,1}]

      \addplot+[mambacolor6] table[x index = 0, y index = 1, header = false, each nth point=100] {expres/dr/CC_mixed_dx6_dz3.dat};
      \addplot+[mambacolor1] table[x index = 0, y index = 1, header = false, each nth point=100] {expres/dr/CC_mixed_dx9_dz3.dat};
      \addplot+[mambacolor3] table[x index = 0, y index = 1, header = false, each nth point=100] {expres/dr/CC_mixed_dx3_dz3.dat};
      \addplot+[mambacolor4] table[x index = 0, y index = 1, header = false, each nth point=100] {expres/dr/CC_mixed_dx1_dz3.dat};
      \addplot[dotted, mark=none, red] coordinates {(0,0.5) (1,0.5)};      
    \end{axis}
  \end{tikzpicture}
  \fi
  \caption[caption]{[Higher is better]
    Accuracy over top-$k\%$ pairs sorted by confidence, for different generating models of $\Xb, Y$, and different dimensionality of $\Xb$, with $\dim(\Zb)=3$ fixed. The baseline is at $0.5$.
    On the left, $\Xb, \Zb$ and $W$ are $\sim \N(0,1)$, while on the right, we consider the adversarial case where we use out-of-model source distributions.
  }
  \label{fig:synth}
\end{figure}
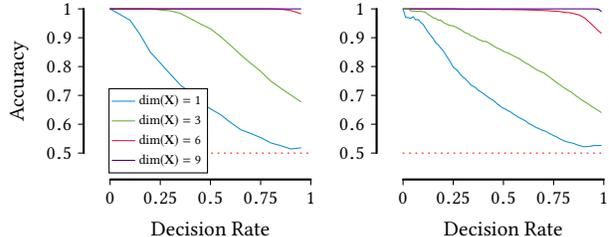

We consider the case where we fix the dimensionality of $\Zb$ to be $3$, and vary the dimensionality of $\Xb$ to be $1,3,6,9$ and further restrict $p_{x}, p_z, p_w$ to be $\Nc(0,1)$, as these are precisely the model assumptions made by \ourmethod.
We show the resulting DR plot in the left plot of Fig.~\ref{fig:synth}. We see that for all dimensionalities of $\Xb$ the pairs for which \ourmethod is most confident are also most likely to be classified correctly.
While for $\dim(\Xb) = 1,3 \leq \dim(\Zb)$ there is (too) little information about $\Zb$ that can be inferred by the factor model, for $\dim(\Xb) = 6, 9 > \dim(\Zb)$ \ourmethod is both highly confident and accurate over all decisions.

Next, we move away from our model assumptions and aggregate over all the possibilities $p_x = p_z = p_w$ listed above. We show the results on the right-hand side of Fig.~\ref{fig:synth}. We observe essentially the same pattern, except that all the lines drop off slightly earlier than in the left plot. Experiments where we chose $p_x, p_z, p_w$ independently at random resulted in similar results and are hence not shown for conciseness. This shows us that our method continues to work even when the assumptions we make no longer hold.

Importantly, all results in both experiments are significant with regard to the $95\%$ confidence interval of a fair coin flip---except for $\dim(\Xb) = 1$ which is significant only for the 75\% of tuples where it was most confident. Further, in none of these cases was the method biased towards classifying datasets as causal or confounded.


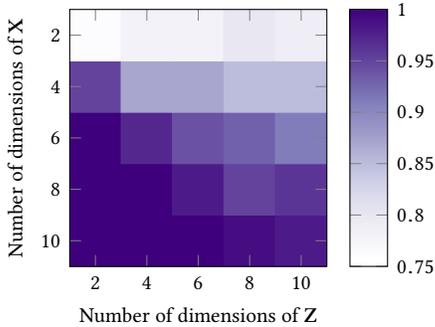
\begin{figure}[t]
\center
  \ifarxiv
  \includegraphics{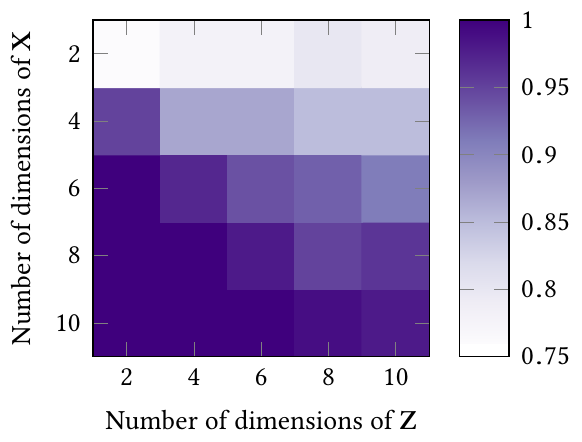}
  \else
  \begin{tikzpicture}
    \begin{axis}[enlargelimits=false,
      colorbar, colormap/Purples,
      colorbar style={ymin=0.75, ymax=1.0, ytick={0.75,0.8,0.85,0.9,0.95,1.0}},
      xlabel=Number of dimensions of $\Zb$,ylabel=Number of dimensions of $\Xb$,
      height=5cm, width=5cm,
      tick label style={font=\footnotesize},
      label style={font=\footnotesize}
      ]
      \addplot [matrix plot,point meta=explicit]
      coordinates {
        (2,10) [1.00] (4,10) [1.00] (6,10) [1.00] (8,10) [0.99] (10,10) [0.98]

        (2,8) [1.00] (4,8) [1.00] (6,8) [0.98] (8,8) [0.95] (10,8) [0.96]

        (2,6) [1.00] (4,6) [0.97] (6,6) [0.94] (8,6) [0.93] (10,6) [0.91]

        (2,4) [0.95] (4,4) [0.87] (6,4) [0.87] (8,4) [0.85] (10,4) [0.85]

        (2,2) [0.76] (4,2) [0.78] (6,2) [0.78] (8,2) [0.80] (10,2) [0.79]
      };
    \end{axis}
  \end{tikzpicture}
  \fi
  \caption{[Darker is better] Area under the Decision Rate Curve heatmap over dimensionality of $\Xb$ and $\Zb$. For fixed $\dim(\Zb)$ performance improves as $\dim(\Xb)$ increases, while for fixed $\dim(\Xb)$ performance degrades as $\dim(\Zb)$ increases. \ourmethod scores between $0.75$ and $1.0$, against a baseline of $0.5$.}
  \label{fig:heat}
\end{figure}

To see how \ourmethod fares for a broader variety of combinations of dimensionalities of $\Xb$ and $\Zb$, in Fig.~\ref{fig:heat} we plot a heatmat of the area under the decision rate curve (AUDR) of \ourmethod. As expected we see that when $\dim(\Zb)$ is fixed we become more accurate as $\dim(\Xb)$ increases. Further as $\dim(\Zb)$ increases for fixed $\dim(\Xb)$ our performance degrades gracefully---this is because we infer a $\hat{\Zb}$ of dimensionality one, which deviates further from the true generating process as the dimensionality of the true $\Zb$ increases. Note that all \ourmethod AUDR scores are above $0.75$, whereas a random classifier would obtain a score of only $0.5$. 

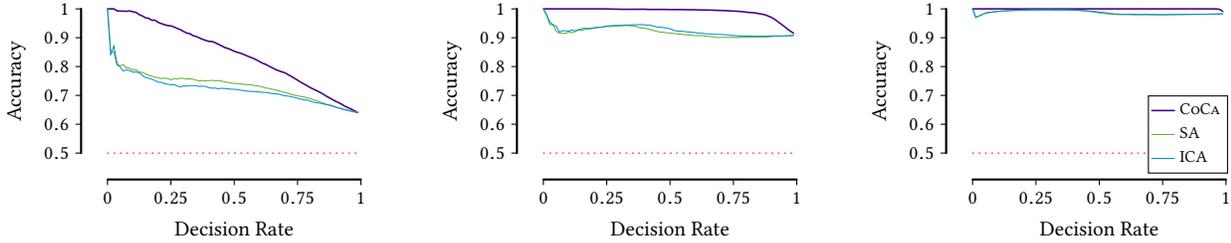
\begin{figure*}[t]
    \begin{minipage}[b]{0.33\linewidth}
      \ifarxiv
      \includegraphics{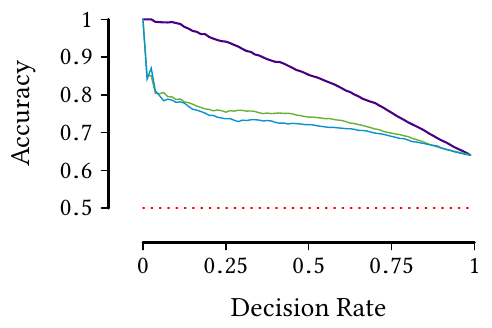}
      \else
      \begin{tikzpicture}
        \begin{axis}[eda line, ymax = 1.0, ymin = 0.5, ytick={1.0,0.9,0.8,0.7,0.6,0.5},
          ylabel={Accuracy}, xlabel={Decision Rate},
          ylabel near ticks, xlabel near ticks,
          tick label style={font=\footnotesize},
          label style={font=\footnotesize},
          height=3.5cm, width=4.95cm,
          legend pos=south east, legend columns=1,
          legend style={nodes={scale=2.0, transform shape}, at={(0.725,0.45)},anchor=north},
          xmin=0, xmax=1, xtick={0,0.25,0.5,0.75,1}]

          \addplot+[mambacolor1, line width=0.6pt] table[x index = 0, y index = 1, header = true, each nth point=100] {expres/dr/CC_mixed_dx3_dz3.dat};
          \addplot+[mambacolor3] table[x index = 0, y index = 1, header = true, each nth point=100] {expres/dr/spectral_mixed_dx3_dz3.dat};
          \addplot+[mambacolor4] table[x index = 0, y index = 1, header = true, each nth point=100] {expres/dr/indep_mixed_dx3_dz3.dat};
          \addplot[dotted, mark=none, red] coordinates {(0,0.5) (1,0.5)};
        \end{axis}
      \end{tikzpicture}
      \fi
     \end{minipage}
	 \begin{minipage}[b]{0.33\linewidth}
      \ifarxiv
      \includegraphics{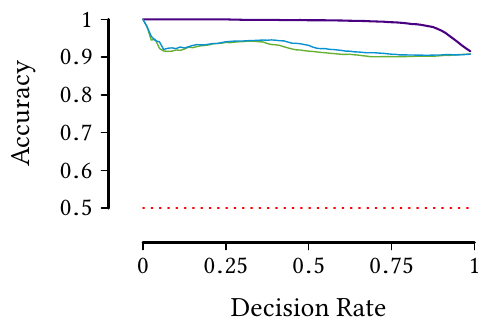}
      \else
      \begin{tikzpicture}
        \begin{axis}[eda line, ymax = 1.0, ymin = 0.5, ytick={1.0,0.9,0.8,0.7,0.6,0.5},
          ylabel={Accuracy}, xlabel={Decision Rate},
          ylabel near ticks, xlabel near ticks,
          tick label style={font=\footnotesize},
          label style={font=\footnotesize},
          height=3.5cm, width=4.95cm,
          legend pos=south east, legend columns=1,
          legend style={nodes={scale=0.5, transform shape}, at={(0.75,0.35)},anchor=north},
          xmin=0, xmax=1, xtick={0,0.25,0.5,0.75,1}]

          \addplot+[mambacolor1, line width=0.6pt] table[x index = 0, y index = 1, header = true, each nth point=100] {expres/dr/CC_mixed_dx6_dz3.dat};
          \addplot+[mambacolor3] table[x index = 0, y index = 1, header = true, each nth point=100] {expres/dr/spectral_mixed_dx6_dz3.dat};
          \addplot table[x index = 0, y index = 1, header = false, each nth point=100] {expres/dr/indep_mixed_dx6_dz3.dat};
          \addplot[dotted, mark=none, red] coordinates {(0,0.5) (1,0.5)};
        \end{axis}
      \end{tikzpicture}
      \fi
     \end{minipage}%
     \begin{minipage}[b]{0.33\linewidth}
      \ifarxiv
      \includegraphics{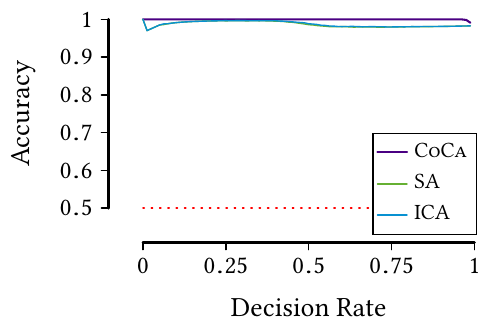}
      \else
      \begin{tikzpicture}
        \begin{axis}[eda line, ymax = 1.0, ymin = 0.5, ytick={1.0,0.9,0.8,0.7,0.6,0.5},
          ylabel={Accuracy}, xlabel={Decision Rate},
          ylabel near ticks, xlabel near ticks,
          tick label style={font=\footnotesize},
          label style={font=\footnotesize},
          height=3.5cm, width=4.95cm,
          legend pos=south east, legend columns=1,
          legend style={nodes={scale=0.9, transform shape}, at={(0.85,0.4)},anchor=north},
          xmin=0, xmax=1, xtick={0,0.25,0.5,0.75,1}]

          \addplot+[mambacolor1, line width=0.6pt] table[x index = 0, y index = 1, header = true, each nth point=100] {expres/dr/CC_mixed_dx9_dz3.dat};
          \addlegendentry{\ourmethod}
          \addplot+[mambacolor3] table[x index = 0, y index = 1, header = true, each nth point=100] {expres/dr/spectral_mixed_dx9_dz3.dat};
          \addlegendentry{SA}
          \addplot+[mambacolor4] table[x index = 0, y index = 1, header = true, each nth point=100] {expres/dr/indep_mixed_dx9_dz3.dat};
          \addlegendentry{ICA}
          \addplot[dotted, mark=none, red] coordinates {(0,0.5) (1,0.5)};
        \end{axis}
      \end{tikzpicture}
      \fi
    \end{minipage}
    \caption[caption]{[Higher is better.]
      Comparing \ourmethod to the spectral~\cite{janzing2017detecting} (SA) and ICA-based~\cite{janzing2018detecting} (ICA) approaches by Janzing and Sch\"{o}lkopf in synthetic data of, from left to right, resp. $\dim(\Xb)=3$, $6$, and $9$. Baseline accuracy is at 0.5. 
      We see that in all cases, \ourmethod performs best by a margin, particularly in regions where it is most confident. 
  }
  \label{fig:comparison}
\end{figure*}

Finally, we compare \ourmethod to the only two competitors we are aware off; the two recent approaches by Janzing and Sch\"{o}lkopf, of which one is based on spectral analysis (SA) \cite{janzing2017detecting} and the other on independent component analysis (ICA) \cite{janzing2018detecting}. The implementation of both methods require $\Xb$ to be multidimensional. We hence consider the cases where $\dim(\Xb) = 3,6,9$, while allowing $p_x, p_y, p_z$ to be any of the distributions listed above. We show the results in Fig.~\ref{fig:comparison}.

As SA and ICA provide an estimate $\widehat{\beta} \in [0, 1]$ measuring the strength of confounding without any two-sided confidence score, we used $| \widehat{\beta} - 1/2 |$ as a substitute for such a score. That the corresponding lines are shaped as expected gives us some assurance that this is a reasonable choice.

We see that for all dimensionalities \ourmethod outperforms these competitors by a margin where the respective methods are most confident, but also that the overall accuracies are almost indistinguishable. We further note that as the dimensionality of $\Xb$ becomes large relative to $\Zb$ the differences in performance between the approaches reduces.

\subsection{Simulated Genetic Networks}
\label{sec:gn}

Next, we consider more realistic synthetic data. For this we consider the DREAM 3 data~\cite{prill2010towards} which was originally used to compare different methods for inferring biological networks. We use this data both because the underlying generative network is known, and because the generative dynamics are biologically plausible \cite{prill2010towards}. That is, the relationships are highly nonlinear, and therefore an interesting case to evaluate how \ourmethod performs when our assumptions do not hold at all. Out of all networks in the dataset, we consider the ten largest networks, those of 50 and 100 nodes, which are associated with time series of lengths 500 and 1000, respectively. Since \ourmethod was not designed to work with time series, we treat the data as if it were generated from an i.i.d. source.

For each network we take pairs $(X, Y)$ of univariate $X$ and $Y$ such that either of the following two cases holds
 \begin{itemize}[noitemsep,topsep=0pt]
\item $X$ has a causal effect on $Y$ and there exists no common parent $Z$, or 
\item $X, Y$ have a common parent $Z$ and there are no causal effects between $X$ and $Y$.
\end{itemize}
Although in theory we could also consider tuples $(X_1,...,X_m, Y)$ with $m > 1$, for this dataset there were too few such tuples to have sufficient statistical power. Further, since the original networks are heavily biased towards causality rather than to common parents we take all the confounded tuples and then uniformly sample an equal number of causal tuples from the set of all such tuples.

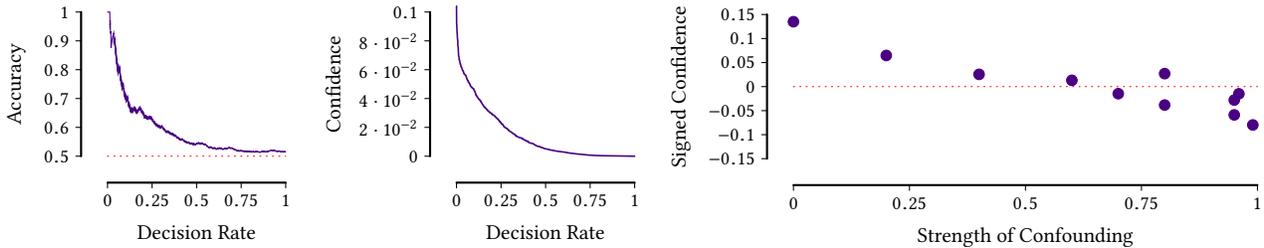
\begin{figure}[t]
  \begin{minipage}[b]{0.5\linewidth}
    \ifarxiv
    \includegraphics{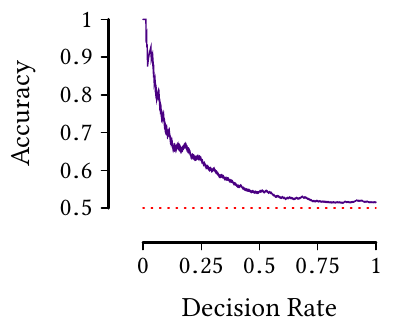}
    \else
    \begin{tikzpicture}
      \begin{axis}[eda line, ymax = 1.0, ymin = 0.5, ytick={1.0,0.9,0.8,0.7,0.6,0.5},
            ylabel={Accuracy}, xlabel={Decision Rate},
            ylabel near ticks, xlabel near ticks,
            tick label style={font=\footnotesize},
            label style={font=\footnotesize},
          height=3.5cm, width=3.95cm,
          legend pos=south east, legend columns=1,
          legend style={nodes={scale=0.5, transform shape}, at={(0.75,0.35)},anchor=north},
          xmin=0, xmax=1, xtick={0,0.25,0.5,0.75,1}]
        \addplot[name path=l, dotted, mark=none, red] coordinates {(0,0.5) (1,0.5)};
        \addplot+[mambacolor1] table[x index = 0, y index = 1, header = false] {expres/gn/dr_gn_2.dat};
      \end{axis}
    \end{tikzpicture}%
    \fi
  \end{minipage}%
  \begin{minipage}[b]{0.5\linewidth}
    \ifarxiv
    \includegraphics{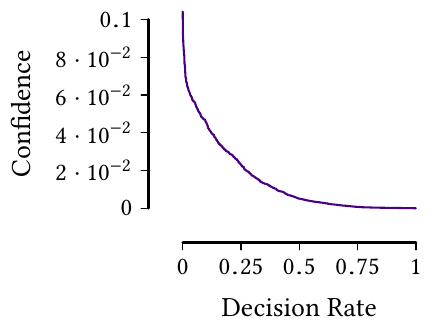}
    \else
    \begin{tikzpicture}
      \begin{axis}[eda line, ymax = 0.10, ymin = 0.0, ytick={0.1,0.08,0.06,0.04,0.02,0.0},
            ylabel={Confidence}, xlabel={Decision Rate},
            ylabel near ticks, xlabel near ticks,
            tick label style={font=\footnotesize},
            label style={font=\footnotesize},
          height=3.5cm, width=3.95cm,
          legend pos=south east, legend columns=1,
          legend style={nodes={scale=0.5, transform shape}, at={(0.75,0.35)},anchor=north},
          xmin=0, xmax=1, xtick={0,0.25,0.5,0.75,1}]
            \addplot+[mambacolor1, line width=0.6pt] table[x index = 0, y index = 1, header = false] {expres/gn/dr_gn_2_conf.dat};
      \end{axis}
    \end{tikzpicture}      
    \fi
  \end{minipage}
  \hfill~
  \caption[caption]{Decision rate and corresponding confidence plots for the genetic networks data.
    \ourmethod is accurate when it is confident, even for this adversarial setting.
  }
  \label{fig:gn}
\end{figure}

We show the decision rate plot when applying \ourmethod to these pairs after aggregating over all the networks in the left-hand side plot of Fig.~\ref{fig:gn}. Like before, we see that \ourmethod is highly accurate for those tuples where it is most confident. In comparison to the results for $\dim(\Xb)=1$ in Fig.~\ref{fig:synth}, we see that performance drops more quickly, which is readily explained by the fact that the simulated dynamics are highly nonlinear. Note however, that our results are nevertheless still statistically significant with regard to a fair coin flip for up the 75\% pairs \ourmethod that is most confident about. To further explain the behavior of \ourmethod on this dataset, we plot the absolute confidence scores we obtain on the right of Fig.~\ref{fig:gn}. We see that particularly for the first ~25\% of the decisions the confidences we obtain are much larger than for the remaining pairs. This corresponds very nicely to the plot on the left, as the first 25\% of our decisions are also those where we compare most favorably to the baseline.



\subsection{T\"{u}bingen Benchmark Pairs}
\label{sec:tb}

To consider real-world data suited for causal inference, we now consider the T\"{u}bingen benchmark pairs dataset.\!\footnote{\url{https://webdav.tuebingen.mpg.de/cause-effect/}}
This dataset consists of (mostly) pairs $(X, Y)$ of univariate variables for which plausible directions of causality can be decided assuming no hidden confounders. For many of these, however, it is either known, or plausible to posit that they are confounded rather than directly causally related. For example, for pairs 65--67 certain stock returns are supposedly causal, but given the nature of the market would likely be better explained by common influences on the returns of the stock options.

We therefore code every pair in the benchmark dataset as either causal (if we think the directly causal part to be stronger), confounded (if we expect the common cause to be the main driver), or unclear (if we are not sure which component is more important) and apply \ourmethod to the pairs in the first two categories. This leaves 47 pairs, of which we judged 41 to be mostly causal and 6 to be mostly confounded.\!\footnote{The complete list can be found in the online appendix at \codeurl.}

\begin{figure}[t]
    \ifarxiv
    \includegraphics{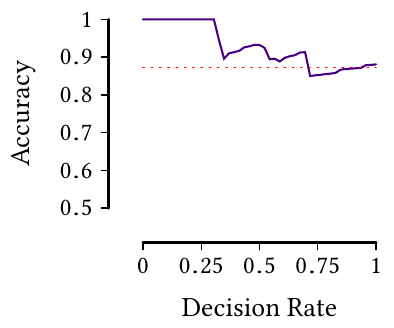}
    \includegraphics{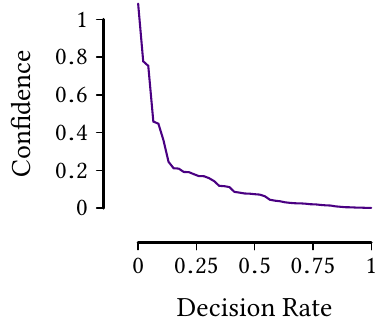}
    \else
    \begin{tikzpicture}
      \begin{axis}[eda line, ymax = 1.0, ymin = 0.5, ytick={0.5,0.6,0.7,0.8,0.9,1.0},
        ylabel={Accuracy}, xlabel={Decision Rate},
        ylabel near ticks, xlabel near ticks,
        tick label style={font=\footnotesize},
        label style={font=\footnotesize},
        height=3.5cm, width=3.95cm,
        legend pos=south east, legend columns=1,
        legend style={nodes={scale=0.5, transform shape}, at={(0.75,0.35)},anchor=north},
        xmin=0, xmax=1, xtick={0,0.25,0.5,0.75,1}]

        \addplot+[mambacolor1, line width=0.6pt]  table[x index = 0, y index = 1, header = false] {expres/pair/dr_pair.dat};
        \addplot[dotted, mark=none, red] coordinates {(0,1-6/47) (1,1-6/47)};
      \end{axis}
    \end{tikzpicture}%
    \begin{tikzpicture}
      \begin{axis}[eda line, ymax = 1.0, ymin = 0.0, ytick={0.0,0.2,0.4,0.6,0.8,1.0},
        ylabel={Confidence}, xlabel={Decision Rate},
        ylabel near ticks, xlabel near ticks,
        tick label style={font=\footnotesize},
        label style={font=\footnotesize},
        height=3.5cm, width=3.95cm,
        legend pos=south east, legend columns=1,
        legend style={nodes={scale=0.5, transform shape}, at={(0.75,0.35)},anchor=north},
        xmin=0, xmax=1, xtick={0,0.25,0.5,0.75,1}]
        \addplot+[mambacolor1, line width=0.6pt] table[x index = 0, y index = 1, header = false] {expres/pair/dr_pair_conf.dat};
      \end{axis}
    \end{tikzpicture}
    \fi
  \hfill~
  \caption[caption]{Decision rate plot and its corresponding confidence plot for the T\"{u}bingen pairs. The baseline for the decision rate plot is at 0.87. Note the strong correspondence between high confidence and high accuracy. 
    }
  \label{fig:pair}
\end{figure}
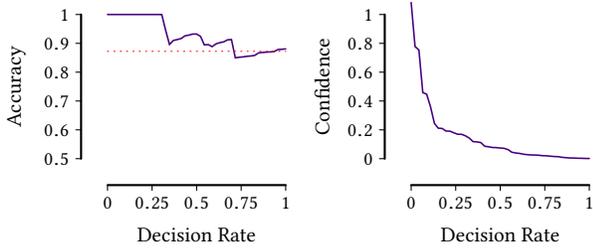

In Fig.~\ref{fig:pair} we show the decision rate plots across the datasets weighed according to the benchmark definition. As in the previous cases, \ourmethod is most accurate where it is most confident, while declining to the baseline as we try to classify points about which \ourmethod is less and less certain. We note that for these cases \ourmethod was biased towards saying that datasets represented truly causal relationships, even when we judged them to be driven by confounding. Despite this, \ourmethod does better than the naive baseline of ``everything is causal'' by assigning more confidence to those datasets which according to our judgment were indeed truly causal.

\subsection{Optical Data}
\label{sec:opt}

Finally, we consider real world optical data~\cite{janzing2017detecting}. In these experiments, $\Xb$ is a low-resolution ($3$$\times$$3$ pixels) image shown on the screen of a laptop and $Y$ is the brightness measured by a photodiode at some distance from the screen. The confounders $\Zb$ are an LED in front of the photodiode and another LED in front of the camera, both controlled by random noise, where the strength of confounding is controlled by the brightness of these LEDs.

\begin{figure}[t]
    \ifarxiv
    \includegraphics{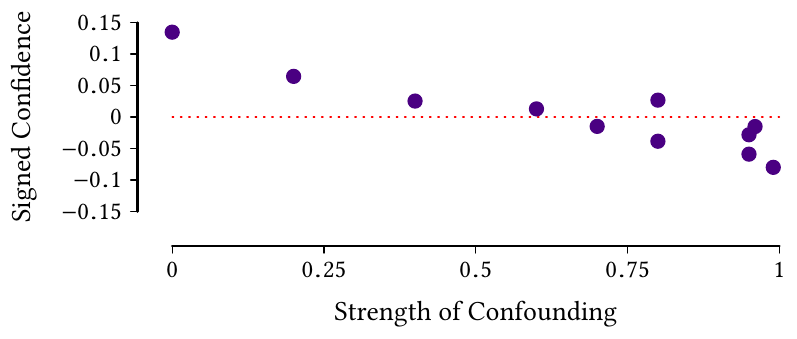}
    \else
    \begin{tikzpicture}
      \begin{axis}[eda line, ymin = -0.15, ymax = 0.15,
        ytick={-0.15, -0.10, -0.05, 0, 0.05, 0.10, 0.15},
        yticklabel style={/pgf/number format/fixed},
        ylabel={Signed Confidence}, xlabel={Strength of Confounding},
        ylabel near ticks, xlabel near ticks,
        tick label style={font=\footnotesize},
        label style={font=\footnotesize},
        height=3.5cm, width=7.75cm,
        legend pos=south east, legend columns=1,
        legend style={nodes={scale=0.5, transform shape}, at={(0.75,0.35)},anchor=north},
        xmin=0, xmax=1, xtick={0,0.25,0.5,0.75,1}]
        \addplot+[only marks, mark=*] table[x index = 0, y index = 1, header = false] {expres/opt/signed_conf.dat};
        \addplot[dotted,red] coordinates {(0,0) (1,0)};
      \end{axis}
    \end{tikzpicture}%
    \fi
  \hfill~
  \caption[caption]{Strength of confounding against the signed confidence of \ourmethod on the optical data. The confounding strength increases from left to right. Higher positive values indicate a stronger belief in causality of \ourmethod, while more negative values indicate a stronger belief in confounding.}
  \label{fig:opt}
\end{figure}

We evaluate \ourmethod on each of the provided datasets, and plot the resulting values in Fig.~\ref{fig:opt}. The strength of confounding increases from the left to right, and values larger than zero indicate that \ourmethod judged the data to be causal, while values smaller than zero indicate confounding. We see that towards an intermediate confounding strength of $0.5$ our method is very uncertain about its classification, while towards the extreme ends of pure causality or pure confounding it is very confident, and correct in being so.



\section{Discussion and Conclusions}
\label{sec:discussion}

We considered the problem of distinguishing between the case where the data $(\Xb, Y)$ has been generated via a genuinely causal model and the case where the apparent cause and effect are in fact confounded by unmeasured variables. We proposed a practical information theoretic way of comparing these cases on the basis of MDL and latent variable models that can be efficiently inferred using variational inference. Through experiments we showed that \ourmethod works well in practice---including in cases where the data generating process is quite different from our models assumptions. Importantly, we showed that \ourmethod is particularly accurate when it is also confident, more so than its competitors.

For future work, we will investigate the behavior of \ourmethod if we use more complex latent variable models, as these allow for modelling more complex relations. These methods, however, also come with a much higher computational cost and without theoretical guarantees of consistency, but may work well in practice. In addition, we would like to be able to infer more complete networks on $(\Xb, Y)$ while taking into account the presence of confounders. However, this will likely lead to inconsistent inference of edges unless we can find a theoretically well-founded method of telling apart direct and indirect effects. To the best of our knowledge, as of now, no such method is known. 


\section*{Acknowledgements}
David Kaltenpoth is supported by the International Max Planck Research School for Computer Science (IMPRS-CS). 
Both authors are supported by the Cluster of Excellence ``Multimodal Computing and Interaction'' within the Excellence Initiative of the German Federal Government.

\bibliographystyle{abbrv}
\bibliography{bib/abbrev,bib/bib-jilles,bib/bib-paper,bib/extracted-bib-david} 

\ifapx
\appendix
\section{Appendix}
\label{sec:apx}

\subsection{Coding of the T\"{u}bingen Pairs}
\label{sec:tub}

Here we give a full list of which pairs of the T\"{u}bingen pairs dataset we considered to be mainly causal, confounded, or which we were uncertain about.

\begin{itemize}[noitemsep,topsep=0pt]
\item Causal: 13--16, 25--37, 43--46, 48, 54, 64, 69, 71--73, 76--80, 84, 86--87, 93, 96--98, 100
\item Confounded: 65--67, 74--75, 99
\item Uncertain: 1--12, 17--24, 38--42, 47, 49--53, 55--63, 68, 70, 81--83, 85, 88--92, 94--95
\end{itemize}

For example for pairs 5--11 it was unclear to us to what extent the age of an abalone should be considered as a causal factor to its length, height, weight, or other measurements, and to what extent all of these should simply be confounded by the underlying biological processes of development.

As another example, for pair 99 we believed that it is reasonable to suggest that the correlation between language test score of a child and socio-economic status of its family might more plausibly be explained by the intelligence of parents and child --- which are strongly correlated themselves.
\fi

\end{document}